\documentclass[aps,twocolumn,reprint,prd,preprintnumbers,nofootinbib]{revtex4-1}
\pdfoutput=1
\usepackage{amsmath,amssymb,amsthm,tikz,tikz-cd,multirow,bm,graphicx,float,array,rotfloat,pgf,soul,bm,xcolor,comment,threeparttable,booktabs}
\usepackage[all]{xy}
\usepackage[shortlabels]{enumitem}
\usepackage[normalem]{ulem}
\usepackage[hidelinks]{hyperref}
\usepackage[vcentermath]{youngtab}
\usepackage[boxsize=.8em]{ytableau}

\begin{document}

\title{Detecting Homeomorphic 3-manifolds via Graph Neural Networks}

\vspace*{-3cm} 
\begin{flushright}
{\tt DESY-24-122}\\
\end{flushright}

\author{Craig Lawrie}
\email[\texttt{craig.lawrie1729@gmail.com}]{}
\affiliation{Deutsches Elektronen-Synchrotron DESY, Notkestr.~85, 22607 Hamburg, Germany}

\author{Lorenzo Mansi}
\email[\texttt{lorenzo.mansi@desy.de}]{}
\affiliation{Deutsches Elektronen-Synchrotron DESY, Notkestr.~85, 22607 Hamburg, Germany}

\begin{abstract}
\noindent 
Motivated by the enumeration of the BPS spectra of certain 3d $\mathcal{N}=2$ supersymmetric quantum field theories, obtained from the compactification of 6d superconformal field theories on three-manifolds, we study the homeomorphism problem for a class of graph-manifolds using Graph Neural Network techniques. Utilizing the JSJ decomposition, a unique representation via a plumbing graph is extracted from a graph-manifold. Homeomorphic graph-manifolds are related via a sequence of von Neumann moves on this graph; the algorithmic application of these moves can determine if two graphs correspond to homeomorphic graph-manifolds in super-polynomial time. However, by employing Graph Neural Networks (GNNs), the same problem can be addressed, at the cost of accuracy, in polynomial time. We build a dataset composed of pairs of plumbing graphs, together with a hidden label encoding whether the pair is homeomorphic. We train and benchmark a variety of network architectures within a supervised learning setting by testing different combinations of two convolutional layers (GEN, GCN, GAT, NNConv), followed by an aggregation layer and a classification layer. We discuss the strengths and weaknesses of the different GNNs for this homeomorphism problem.
\end{abstract}

\maketitle  

\section{Introduction}\label{sec:Intro}

At least since the discovery that expectation values of Wilson loops in Chern--Simons quantum field theory reproduce the Jones polynomial of knots and links \cite{Witten:1988hf}, and the associated connection with the Witten--Reshetikhin--Turaev (WRT) invariant of closed oriented three-manifolds \cite{MR1091619}, low dimensional topology has provided inspiration for high-energy physics. In particular, we can consider 6d maximally-supersymmetric quantum field theories \cite{Witten:1995gx}, and compactify three of the six dimensions on a compact three-manifold, $M_3$. This procedure can lead to an effective 3d $\mathcal{N}=2$ supersymmetric field theory for which the physical properties depend on the topological properties of $M_3$. It has been proposed in \cite{Gukov:2016gkn, Gukov:2017kmk} that the BPS spectrum of this resulting 3d theory, on a certain background, forms a $q$-series (graded by the ``energy'' of the states), referred to as $\widehat{Z}(q)$, which is related to a categorification of the WRT invariant of $M_3$. However, much remains to be discovered about the $\widehat{Z}$-invariants, in particular, a rigorous mathematical definition for general 3-manifolds is yet to be found. As a topological invariant, the $q$-series capturing the BPS spectra must be identical for homeomorphic $M_3$; this motivates us to study the homeomorphism problem for a particular class of three-manifolds where the $\widehat{Z}$-invariants are concretely defined.

It is well-known \cite{MR151948,MR125588} that every irreducible, closed, orientable three-manifold can be encoded in the planar diagram of a link, together with the assignation of integer framing data to each link component. The three-manifold itself is obtained via the application of Dehn surgery to this framed link. Two such three-manifolds are homeomorphic if and only if the underlying framed links are equivalent modulo a sequence of transformations known as the Kirby moves. Since every topological three-manifold admits a unique smooth structure, such a sequence of Kirby moves equivalently specifies a diffeomorphism between smooth three-manifolds.

Determining the existence of a homeomorphism between a given pair of three-manifolds associated with framed link diagrams, $L_1$ and $L_2$, can then be stated as a simple game: the player performs Kirby moves on $L_1$ until the transformed link reproduces $L_2$. The winning player either completes this task or else decides the task is impossible and retires, in the minimal number of moves.

When the state space of a game is large (or even infinite, as it is in our homeomorphic three-manifold game) but the number of possible moves with which one can walk in that state space is small, machine learning algorithms have demonstrated great success in finding a path through the state space to a desired state. This is exactly the situation that holds for board games such as Go and Chess; to such games, a reinforcement approach has been effectively applied \cite{DBLP:journals/corr/abs-1911-08265}. In fact, we may have humbler aims; while outputting the path through state space would have the benefit of being mathematically verifiable, we can instead simply ask the algorithm to output whether it thinks that the manifolds are homeomorphic --  that is, whether a path exists. The latter formulation lies in the domain of supervised learning. 

A possible first step to understanding which three-manifolds are homeomorphic is to understand when two links are isotopic. Each link can be represented by a link diagram and two links are isotopic if and only if their link diagrams are related by a sequence of the three Reidemeister moves \cite{MR3069462,MR1502807}, which form a subset of the Kirby moves. The naive application of these rules results in a super-polynomial time complexity for finding isotopies; therefore, a machine learning approach can be used to produce an approximate solution in polynomial time. Such techniques have been used to study the isotopy of knots via Reidemeister moves in \cite{Gukov:2020qaj,Gukov:2024buj}; machine learning techniques have been applied more generally to problems in knot theory in a number of articles, e.g., \cite{Craven:2020bdz,Craven:2021ckk,Jejjala:2019kio,MR4101599,Davies2021AdvancingMB}. Moving beyond the realm of knots, in this article we focus on a special class of 3-manifolds for which the Kirby diagrams/Kirby moves simplify.

This is, in fact, just one of the many applications that machine learning has found in the context of string theory/algebraic topology/algebraic geometry: the reviews \cite{He:2023csq,Bao:2022rup,Dechant:2022ccf,Ruehle:2020jrk} illustrate how artificial intelligence (AI) can be used in physics to (re)discover known results and make further predictions that were hindered by a lack of computational power, whereas in \cite{Ruehle:2020jrk,Larfors:2021pbb} a more systematic AI-driven approach is used to sample string vacua and derive non-trivial ``experimental'' relations between characterizing quantities. Lastly, more along the direction we pursue in this paper, in \cite{Gukov:2020qaj,Gukov:2024buj,Ri:2023xcn}, machine learning is used in knot theory to probe mathematical conjectures regarding 4d topology or via reinforced learning approaches to produce results in 3d topology that can be hand-checked by a human.

An important subclass of the broad landscape of three-manifolds is formed of those spaces known as graph-manifolds.\footnote{This subclass of three-manifolds are also those for which, in some cases, an explicit formula for computing the $\widehat{Z}$-invariants exist \cite{Gukov:2017kmk}.} Such a space possesses a decomposition, known as the JSJ decomposition \cite{Jaco_Shalen_1979,Johannson_1979}, where the space is cut along embedded tori into a collection of 
\begin{equation}\label{eqn:cuts}
    F_i \times S^1 \,,
\end{equation}
where the $F_i$ are bounded compact two-manifolds, and the three-manifold is constructed via glueing along the boundaries. Any JSJ decomposition can be captured by a graph, $\Gamma$, known as a plumbing graph; this graph is then sufficient to capture the topology of the three-manifold, which we refer to schematically as $M(\Gamma)$. Given two such graph-manifolds, $M(\Gamma_1)$ and $M(\Gamma_2)$, it has been shown that the two manifolds are homeomorphic if and only if the plumbing graphs $\Gamma_1$ and $\Gamma_2$ are equivalent under an equivalence relation known as the von Neumann moves \cite{Neumann_1981}. That is, the Kirby moves on the associated link diagrams reduces to the von Neumann moves on the plumbing graph.

The von Neumann moves describe an algorithm to determined when two graph-manifolds are homeomorphic: recursively perform the von Neumann moves (described in Section \ref{sec:3-manifold}) on $\Gamma_1$ until $\Gamma_1 \cong \Gamma_2$ where $\cong$ denotes the standard graph isomorphism. Unfortunately, such an algorithm takes super-polynomial time in the size of the graph,
rendering it unfeasible to distinguish all but the simplest graph-manifolds.\footnote{See 
\cite{MR4007377} for an analysis of the computational complexity of the homeomorphism problem for three-manifolds.} This is a similar drawback that we observed when discussing the finding of link isotopies, and the success of \cite{Gukov:2020qaj,Gukov:2024buj} encourages us to take a machine learning approach to graph-manifolds. In fact, for a subclass of graph-manifolds, those where the plumbing graph is a tree, the genus of all Riemann surfaces in equation \eqref{eqn:cuts} are zero, and where only the subset of the von Neumann moves preserving that structure are implemented, such a machine learning approach has been successfully pioneered in \cite{Ri:2023xcn}.

As opposed to framed link diagrams, which lie within the purview of mathematics, graphs are a ubiquitous data structure across the fields of human endeavor. For example, a social networking service may depend on a graph where the nodes represent users and the edges capture relationships between those users; the analysis of this graph (for example, for the purpose of maximizing advertising revenue) is clearly of prime importance. Consequently, many tools have been developed that attempt to draw useful conclusions from graphs. One family of such tools are Graph Neural Networks (GNNs), which take advantage of deep learning based methods designed to exploit the graph structure of their domain; see \cite{Zhou_2021} for a review.

In this paper, we construct GNNs, that take as input two graphs, $\Gamma_1$ and $\Gamma_2$, and return a yes/no answer to the question of whether the three-manifolds $M(\Gamma_1)$ and $M(\Gamma_2)$ are homeomorphic, in polynomial time. We use supervised learning to train GNNs, with a variety of different architectures of the convolutional layers, and evaluate the accuracy of the different GNNs. 

An important first step in this direction was taken in \cite{Ri:2023xcn}, where the authors used GNNs to study the homeomorphic equivalence of closed unoriented acyclic graph-manifolds whose JSJ decomposition involves only genus zero Riemann surfaces. This paper therefore forms an intermediate step between the analysis of \cite{Ri:2023xcn} and the implementation of the full state space of Kirby diagrams related via Kirby moves. In contrast to \cite{Ri:2023xcn}, in this paper, we do not use reinforcement learning, where the GNN is trained to give as output a sequence of von Neumann moves; such an output can be rigorously checked by hand, in polynomial time. This would be an exciting future direction which goes beyond our black-box supervised learning approach. 

While an ideal and perfectly trained network can be regarded as a set of topological invariants that distinguish homeomorphic graph-manifolds, any realistic network can only approximate this behavior. Instead, one should interpret the trained GNN as a signpost pointing to interesting graph-manifolds that deserve a subsequent analytical exploration.

\section{Plumbed 3-manifolds}\label{sec:3-manifold}

As discussed in Section \ref{sec:Intro}, an interesting subclass of irreducible three-manifolds are those which admit a JSJ decomposition and thus for which the topology can be entirely captured by a plumbing graph.\footnote{We note that these three-manifolds may have boundary and may be unorientable.} The plumbing graph is a connected undirected graph (possibly with multiple edges between any pair of vertices and non-trivial cycles) where the vertices and edges carry labels.\footnote{An extension to disconnected graphs is also possible, corresponding to connected sums of three-manifolds. We do not explore such cases in this paper.}

In this section, we focus only on elucidating the structure of such graphs and the behavior of the graphs under the von Neumann moves; this is sufficient to set up the problem to which we apply the Graph Neural Networks. We encourage the reader to consult \cite{Neumann_1981}, of which we adopted both conventions and notation, for a detailed description of the topology associated with each of the graph features and the von Neumann moves.

\begin{figure}[t]
    \centering
    \includegraphics[page=1,scale=1.2]{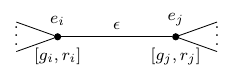}
    \caption{Example of plumbing graph $\Gamma$ with the triplet $(e_i,g_i,r_i)$ explicitly written at the vertices and an $\epsilon=\pm 1$ oriented edge.}
    \label{fig:plumbing_dictionary}
\end{figure}

Each vertex in the plumbing graph corresponds to an $F_i \times S^1$ in the JSJ decomposition and is labelled by a triplet of integers
\begin{equation}\label{eqn:XXX}
    (e_i, g_i, r_i \geq 0) \,.
\end{equation}
Here $e_i$ and $g_i$ are, respectively, the Euler number and the genus of the $F_i$,\footnote{We use the convention that $g_i < 0$ corresponds to non-orientable surfaces.} and $r_i$ counts the number of open disks removed from $F_i$; if $r_i > 0$, then $e_i$ must vanish.\footnote{For ease of notation, when writing the von Neumann moves sometimes there would be a non-trivial $e_i$ associated with a vertex with $r_i > 0$. In these cases, the label $e_i$ should be disregarded.} Associated with each edge of the plumbing graph is a sign
\begin{equation}
    \epsilon = \pm 1 \,,
\end{equation}
which is referred to as orientation. We also define the degree of a vertex, $d_i$, as the number of edges incident to that vertex. Each $F_i$ contains $d_i + r_i$ boundary components, of which $d_i$ are glued to another $F_j \times S^1$ in the JSJ decomposition. We have depicted a fragment of the plumbing graph in Figure \ref{fig:plumbing_dictionary}. When any of the triplet $(e_i, g_i, r_i)$ is zero, we often do not write that quantity explicitly in the graph.

Given such a plumbing graph $\Gamma$, the graph-manifold, $M(\Gamma)$, is schematically obtained as follows. For each vertex in $\Gamma$, let $F_i$ be a compact two-manifold of genus $g_i$ with $d_i + r_i$ boundary components. Let $E_i$ be an oriented circle-bundle $E_i \rightarrow F_i$ with total space $E_i$, and with a chosen trivialization $E_i |_{\partial F_i}$, such that the Euler number is $e_i$. If $g_i \geq 0$, we choose an orientation of $F_i$ and the trivialization in a compatible manner. If vertices labelled by $i$ and $j$ are connected with an edge labelled by orientation $\epsilon$ then a boundary component $S^1 \times S^1$ of $E_i$ is glued to a boundary component $S^1 \times S^1$ of $E_j$ via the map
\begin{equation}
    \epsilon J = \epsilon \begin{pmatrix} 0 & 1 \\ 1 & 0 \end{pmatrix} \, : \,\, S^1 \times S^1 \rightarrow S^1 \times S^1 \,.
\end{equation}
We have used that for any $A = \left(\begin{smallmatrix} a & b \\ c & d \end{smallmatrix}\right) \in SL(2, \mathbb{Z})$ there is an associated torus diffeomorphism
\begin{equation}
    \begin{aligned}
        A : \, S^1 \times S^1 &\rightarrow S^1 \times S^1 \\
        (t_1, t_2) &\mapsto (t_1^at_2^b, t_1^ct_2^d) \,.
    \end{aligned}
\end{equation}

For the subclass of graph-manifolds studied via supervised learning techniques in \cite{Ri:2023xcn}, each vertex is parametrized by a single integer, and the edge labels are all $+1$; such a graph can be captured by an adjacency matrix where the diagonal entries are the integers associated with the vertices. 
\subsection{Homeomorphic Graph-manifolds}

Given a plumbing graph $\Gamma$, with its associated graph-manifold $M(\Gamma)$, it is known how to construct three-manifolds homoeomorphic to $M(\Gamma)$ by acting on $\Gamma$ via a series of moves that preserve the oriented diffeomorphism type of $M(\Gamma)$. As described in \cite{Neumann_1981}, these moves form a semi-group and are referred to by the labels R0 through R8. Furthermore, any homeomorphism $M(\Gamma_1) \cong M(\Gamma_2)$ can be written as a sequence of these moves and their inverses. We implement and review a subset of the moves, those which do not change the number of connected components of the graph, here.

\begin{itemize}
    \item \textbf{\textit{$\mathrm{R0}$ -- Reversal}}: if a vertex has $g\ge0$ reverse the orientation of all the edges connected to it other than loops; else if $g<0$, reverse also the loops.
    \item \textbf{\textit{$\mathrm{R1}$ -- Blowing down}}: replace one of the subgraphs on the left with the corresponding subgraph on the right. Here $\rho=\pm 1$ and $\epsilon_0=-\rho \epsilon_1 \epsilon_2$.
    \[
    \includegraphics[page=2,width=0.9\columnwidth]{ML_Plumbed_Manifold_Figures.pdf}
    \]
    \item \textbf{\textit{$\mathrm{R2}$ -- $\mathbb{RP}^2$ absorption}}: replace the subgraph on the left with the one on the right. Here $\delta_1=\pm 1$, $\delta_2=\pm 1$, and $\delta=\frac{\delta_1+\delta_2}{2}$. 
        \[
    \includegraphics[page=3,width=0.9\columnwidth]{ML_Plumbed_Manifold_Figures.pdf}
    \]
    \item \textbf{\textit{$\mathrm{R3}$ -- 0-chain absorption}}: replace the subgraph on the left with the one on the right. The orientation of the edges in the right subgraph is given by $\epsilon'_i=-\epsilon \Bar{\epsilon} \epsilon_i$ if the edge is not a loop, otherwise it stays unchanged.
        \[
    \includegraphics[page=4,width=0.9\columnwidth]{ML_Plumbed_Manifold_Figures.pdf}
    \]
    \item \textbf{\textit{$\mathrm{R4}$ -- Unoriented handle absorption}}: replace the subgraph on the left with the one on the right.
        \[
    \includegraphics[page=5,width=0.9\columnwidth]{ML_Plumbed_Manifold_Figures.pdf}
    \]
    \item \textbf{\textit{$\mathrm{R5}$ -- Oriented handle absorption}}: replace the subgraph on the left with the one on the right.
        \[
    \includegraphics[page=6,width=0.9\columnwidth]{ML_Plumbed_Manifold_Figures.pdf}
    \]
    \item \textbf{\textit{$\mathrm{R8}$ -- Annulus absorption}}: replace the subgraph on the left with the one on the right.
        \[
    \includegraphics[page=7,width=0.9\columnwidth]{ML_Plumbed_Manifold_Figures.pdf}
    \]
\end{itemize}
Only when we write a double arrow the inverse move is implemented. For processes that involve a modification of the genus, the operation $\#:\mathbb{Z} \times \mathbb{Z} \rightarrow \mathbb{Z}$ induces a semi-group structure:
\begin{equation}
    g_1 \# g_2 = \begin{cases}
        g_1+g_2 & \text{if }g_1 g_2 \ge 0 \,, \\
        -2g_1+g_2 & \text{if }g_1>0 , g_2<0 \,, \\
        g_1-2g_2 & \text{if } g_1<0 , g_2 > 0 \,.
    \end{cases}
\end{equation}

\section{GNN-Equivalent Graphs}\label{sec:GNN}

Now that we have discussed, in Section \ref{sec:3-manifold}, the structure of the plumbing graphs and the transformations under which they are considered equivalent, we are ready to construct Graph Neural Networks that attempt to determine when two such graphs are equivalent. 

A Graph Neural Network is generically made up of three modules: propagation, sampling, and pooling. Since we are dealing with \textit{small} undirected homogeneous static graphs,\footnote{We define small graphs as those whose adjacency matrix can be fully stored on a machine.} the only module that plays a role in the neural network structure is the propagation module. For the more general architecture of GNNs, as well as for a detailed explanation of the notation used in this section, we refer the reader to the extensive reviews \cite{Zhou_2021, Khemani_2024} and references therein. For the standard ML terminology in this section, we refer to \cite{Alzubaidi2021ReviewOD} for an extensive summary.

The main building blocks inside the propagation module are the convolutional layers. Such a layer can be based on the spectral Fourier transform of the graph, and is thus referred to as Spectral, or else it can be Spatial, in which case it is directly based on the topology of the graph. We consider propagation modules involving two convolutional layers, including both those of Spectral (GCN) and Spatial (GAT, GEN, NNConv) types. Of these four different types of convolutional layers, only GAT, GEN, and NNConv are designed to incorporate the data associated with the edges of the graph; however, utilizing GCN as well allows us to determine the sensitivity of the existence of homeomorphisms on the orientations of the edges.

The GNNs that we consider consist of two convolutional layers, an aggregator layer, and finally a classification layer.\footnote{The code is available on \href{https://github.com/Lollo0900/Plumbed_3-Manifolds}{\color{blue}$\text{GitHub}^{\includegraphics[scale=0.07]{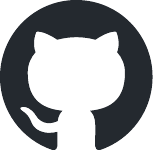}}$}.} The first convolutional layer takes as input a single graph and transforms the data associated with each node, schematically from $\mathbf{x}_i$ to $\mathbf{x}_i'$; the particular choice of convolutional layer fixes the transformation, which depends on the edges in the input graph. The second convolutional layer further transforms the $\mathbf{x}_i'$. The output of the second convolutional layer is passed to the aggregation layer, which determines the graph embedding by combining the features associated with each node coming from the second convolution. Given its performance in \cite{Ri:2023xcn} we utilize the same aggregator, which was first defined in \cite{Li_2015}:
\[
\mathbf{h}_G =\mathrm{MLP} \left( \sum \limits_{i \in V} \mathrm{Softmax}(\mathrm{MLP_{\text{gate}} (\mathbf{x}_i)})\odot \mathrm{MLP}(\mathbf{x}_i)  \right) \,.
\]
Here, $\mathbf{h}_G$ denotes a graph embedding and $\odot$ element-wise multiplication. Finally, there is the classification layer, which takes as input the concatenation of the output of the aggregation layer applied to two different graphs and returns a binary output, i.e., whether the network thinks the graphs correspond to homeomorphic 3-manifolds or not. In Table \ref{tab:NNScheme}, we depict a schematic representation of all the different architectures analysed.

\begin{table*}[t]
    {\renewcommand{\arraystretch}{1.5}
    \begin{tabular}{|>{\centering\arraybackslash} m{0.12\textwidth}|>{\centering\arraybackslash} m{0.12\textwidth}|>{\centering\arraybackslash} m{0.12\textwidth}|>
    {\centering\arraybackslash} m{0.13\textwidth}|>
    {\centering\arraybackslash} m{0.14\textwidth}|>{\centering\arraybackslash} m{0.14\textwidth}|>{\centering\arraybackslash} m{0.18\textwidth}|}        
        \hline
        Layers  & GEN+GAT & GCN+GCN  & GCN+GAT & NNConv+GCN  & NNConv+GAT & NNConv+NNConv \\ \hline
        First convolution & GEN(3,64,1) & \multicolumn{2}{c|}{GCN(3,64)} &  \multicolumn{3}{c|}{NNConv(3,64,1)}\\
        \hline
        Second convolution & GAT(64,64,1) & GCN(64,64) & GAT(64,64,1) & GCN(64,64) & GAT(64,64,1) &  NNConv(64,3,1)\\
        \hline
        Aggregation& \multicolumn{6}{c|}{Aggregator(64,32)}\\
        \hline
        Classification & \multicolumn{6}{c|}{MLP(64,2)}\\
        \hline
    \end{tabular}}
        \centering
    \caption{The architecture of the six GNN models implemented together with their parameters' values.}
    \label{tab:NNScheme}
\end{table*}

In the remainder of this section, we briefly review the architecture of the different convolutional layers herein employed.

\subsection{Graph Embedding Network (GEN)}

A Graph Embedding Network \cite{Li_2019}, designed for graph similarity problems, embeds each graph in a vector referred to as a graph embedding. The structure of the network involves an encoder and a propagator. The encoder maps the node data $\mathbf{x}_i$ and the edge attribute $e_{i,j}$ via:
\begin{equation}
    \begin{aligned}
        \mathbf{x}_i^{(0)} &= \mathrm{MLP}_{\text{node}}(\mathbf{x}_i) \,, \\
            \widetilde{e}_{ij} &= \mathrm{MLP}_{\text{edge}}(e_{i,j}) \,.
    \end{aligned}
\end{equation}
The propagator defines the message
\begin{equation}
    m_{i\rightarrow j}^{(n)} = \mathrm{MLP}\left(\mathbf{x}_i^{(n)},\mathbf{x}_j^{(n)},\tilde{e}_{ij}\right) \,,
\end{equation}
which is then used recursively to update the node data of the graph embedding via single message propagation:
\begin{equation}
    \mathbf{x}_i^{(n+1)}=\mathrm{MLP}\left(\mathbf{x}_i^{(n)}, \sum \limits_{j \in \mathcal{N}(i)} m^{(n)}_{i\rightarrow j}  \right) \,.
\end{equation}
Here, $\mathcal{N}(i)$, known as the local neighborhood of $i$, denotes the set of node indices with non-zero edges to the $i$th node. The GEN layer requires the specification of a discrete parameter: the number of times to propagate the message. In this paper, we propagate the message only once, and thus the output from the GEN layer is $\mathbf{x}_i' = \mathbf{x}_i^{(1)}$.

\subsection{Graph Convolutional Network (GCN)}

The second class of convolutional layer that we consider is known as a Graph Convolutional Network, defined in \cite{Kipf_2017}. Let $E_{j,i}$ denote the edge weight between nodes $i$ and $j$, and we define the degree of the $i$th node as
\begin{equation}
    \widehat{d}_i=1+\sum \limits_{j\in \mathcal{N}(i)} E_{j, i} \,.
\end{equation}
Then, the GCN layer acts on the node parameters via
\begin{equation}
    \mathbf{x}'_i=\mathbf{\Theta}^\intercal\sum_{j\in \mathcal{N}(i)\cup \{i\}}\frac{E_{j,i}}{\sqrt{\widehat{d}_j\widehat{d}_i}}\mathbf{x}_j \,,
\end{equation}
where $\mathbf{\Theta}$ is a matrix of filter parameters. 

\subsection{Graph Attention Network (GAT)}

A Graph Attention Network, defined in \cite{Velickovic_2017}, incorporates the so-called attention mechanism into the message propagation. We define the attention coefficients via
\begin{equation}
    \alpha_{i,j}=
            \frac{\beta_{i,j}}{\sum \limits_{k\in\mathcal{N}(i)\cup\{i\}}\beta_{i,k}}\,,
\end{equation}
where
\begin{equation}
  \begin{aligned}
    \beta_{i,\ell} &= \exp\left(\mathrm{LeakyReLU}(\mathbf{a}_x^\intercal[\boldsymbol{\rho}_{i,\ell}] + \mathbf{a}_e^\intercal \mathbf{\Theta}_e e_{i,\ell})\right) \,, \\
    \boldsymbol{\rho}_{l,n} &= \mathbf{\Theta}_x\mathbf{x}_l+ \mathbf{\Theta}_x\mathbf{x}_n \,.
  \end{aligned}
\end{equation}
The attention mechanisms, denoted as $\mathbf{a}_x$ and $\mathbf{a}_e$, are implemented via single-layer feed-forward neural networks (with filter parameters $\mathbf{\Theta}_x$ and $\mathbf{\Theta}_e$) for the node features $\mathbf{x}_i$ and the edge attribute $e_{i,j}$. Then, the node features are updated by the convolutional layer as follows:
\begin{equation}
    \mathbf{x}'_i = \alpha_{i,i}\mathbf{\Theta}\mathbf{x}_i+\sum_{j\in\mathcal{N}(i)}\alpha_{i,j}\mathbf{\Theta}\mathbf{x}_j \,,
\end{equation}
where $\Theta$ is, as usual, the filter parameters.

\subsection{Edge Conditioned Convolution (NNConv)}

The final type of convolution layer that we consider is an Edge Conditioned Convolution network \cite{Gilmer_2017}. This involves updating the node data utilizing a neural network, $h_{\mathbf{\Theta}}$, for the edge attributes; in our modes this network is constructed from three layers: a Linear layer, a ReakyReLU layer, and another Linear layer. The node data is then updated as follows:
\begin{equation}
    \mathbf{x}'_i=\mathbf{\Theta}\mathbf{x}_i + \sum \limits_{j \in \mathcal{N}(i)} \mathbf{x}_j \cdot h_{\mathbf{\Theta}}\left( e_{i,j} \right) \,.
\end{equation}

\section{Results}\label{sec:results}

\begin{figure*}[t]
    \includegraphics[width=0.7\textwidth]{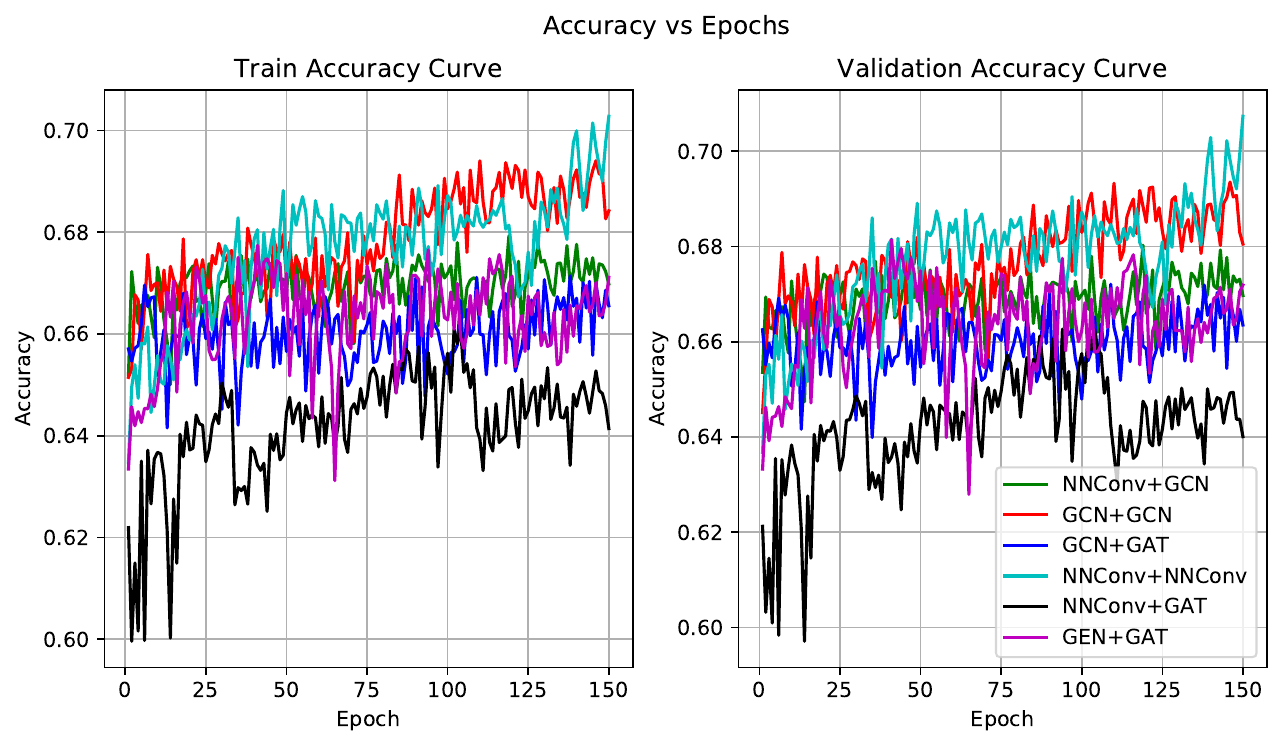}
    \caption{Accuracy curve over the number of epochs for the studied GNN models both for the training set, on the left image, and for the validation one on the right.}
    \label{fig:accuracy}
\end{figure*}

In Section \ref{sec:3-manifold}, we have summarized the structure of plumbing graphs associated with graph-manifolds, and enumerate the von Neumann moves which define equivalence classes of such plumbing graphs. These equivalence classes are precisely the homeomorphism-classes of the associated graph-manifolds. In Section \ref{sec:GNN}, we introduced certain Graph Neural Networks designed to take as input a pair of plumbing graphs and output a yes/no on whether the pair represents homeomorphic graph-manifolds. In this section, we train each of the GNNs discussed in Section \ref{sec:GNN}, and analyze and compare how successful the different architectures are at predicting the existence of homeomorphisms between graph-manifolds. 

For each of the various models, we trained on a dataset consisting of $80000$ pairs of plumbing graphs. Of these pairs, $40000$ were generated using \texttt{EquivPair}, guaranteeing that they correspond to homeomorphic graph-manifolds, and the remaining $40000$ pairs were generated by an equal split between \texttt{InEquivPair} and \texttt{TweakPair}, and were marked as non-homeomorphic graph-manifolds. For details of these three algorithms, see Appendix \ref{app:Algorithms}. We performed a training/validation split of $80\%/20\%$ and then trained the networks via an Adam optimiser with a learning rate of $5 \times 10^{-3}$. The training was run for $150$ epochs per model and was benchmarked by a Cross-Entropy loss function. 

In Figure \ref{fig:accuracy}, we have collected the dependence of the accuracy on the number of epochs on the split dataset for each of the studied GNNs. The GNNs are labelled by the architecture of the two convolutional layers. The average accuracy, together with the standard deviation on the validation sets, truncated at the first digit, is given in Table \ref{tbl:accuracy}. The error for each architecture has been determined by doing statistics on the performance of the last-saved most-optimal tuning in the following training epochs. For the NNConv+NNConv GNN, which reached the most optimal tuning in the final (150th) epoch, we have placed a conservative error of $0.4$ based on the previous performance of this model across multiple trainings.

\begin{figure}[t]
    \centering
    \includegraphics[width=0.9\linewidth]{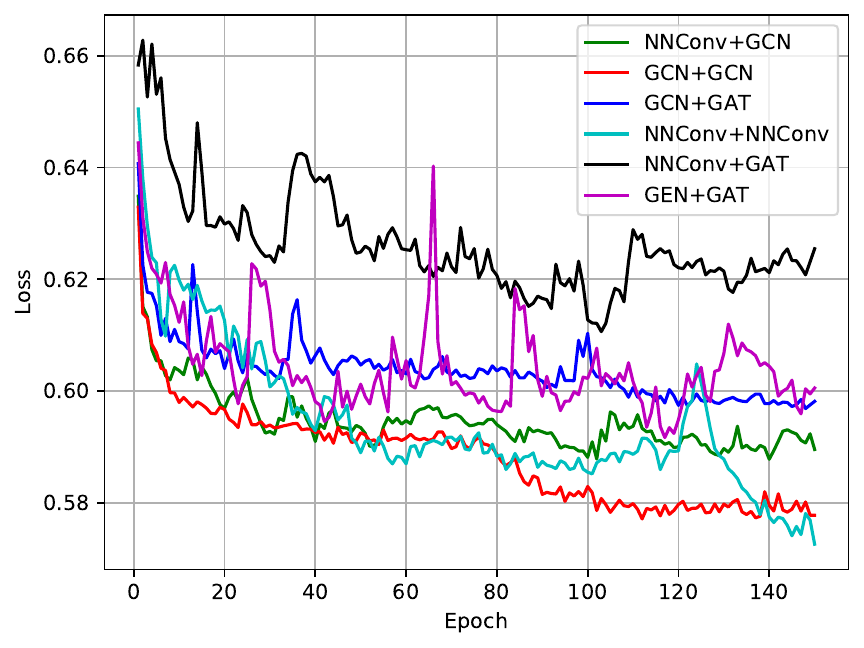}
    \caption{Plot of the Cross-Entropy loss function versus the epochs for each of the studied GNN models. Notice that the loss function for the NNConv+NNConv GNN has not stabilized at the end of the training run.}
    \label{fig:loss}
\end{figure}

Interestingly, the best-performing GNNs are those where the convolutional layers have the same architecture (NNConv+NNConv and GCN+GCN). It can be appreciated that the combination of two graph convolutional layers of GCN or NNConv type works, in general, better than a mixed two layer model with a graph convolution layer plus a layer of graph attention or graph embedding type. In fact, we notice that the maximal value of $70.7\%$ accuracy for the NNConv+NNConv model was attained at the very end of the training run. Hence, a longer training time can, in principle, provide even better results at the cost of training the network with a larger database (with the current $80000$ pair dataset, the training time is $\sim 5$ days). This is also expected since the loss function (shown in Figure \ref{fig:loss}) still demonstrates an improving margin that is not fully exploited due to the limit on the number of training epochs. We note also that the aggregation layer of our GNNs does not involve the edge labels; a more sophisticated aggregator, that takes into account the edge labels, may further improve the accuracy.

Recall that the GCN layer, as opposed to the NNConv layer, is insensitive to the edge labels; however, this does not appear to lead to a significant variation in the accuracy of the GCN+GCN and NNConv+NNConv architectures. This may be a consequence of the R0 von Neumann move, which demonstrates that the homeomorphism class depends only weakly on the edge labels: for isomorphic graphs with the same node data $(e_i, g_i, r_i)$, many of the consistent assignations of $\pm 1$ edge labels are related by the R0 move. To compare the sensitivity of the GNNs to the edge labels, it would be interesting to keep track of the importance of the edge labels in the dataset; that is, to determine what percentage of pairs in the dataset are identical modulo the edge labels, and then to quantify to what extent the accuracy difference between GCN+GCN and NNConv+NNConv is due to such pairs. We leave this for a future analysis.

In contrast to the study in \cite{Ri:2023xcn}, which focused only on a restricted subset of the graph-manifolds and von Neumann moves explored here, the GEN+GAT model does not outperform the other combinations of convolutional layers. Based on the previous paragraph, we do not expect that this is due to the addition of edge labels (which are irrelevant in \cite{Ri:2023xcn}); perhaps such convolutional layers perform with higher efficiency on trees (i.e., acyclic graphs) and this leads to the comparatively poorer performance here. The GCN+GAT and GCN+GCN architectures perform only marginally worse than in \cite{Ri:2023xcn}, despite the significantly more complex features of the dataset studied here. The value of the loss function for each of the models implemented versus the training epochs is plotted in Figure \ref{fig:loss}, and this highlights the aforementioned comparative evaluations of the various GNNs studied.

\begin{table}[t]
    \centering
    \begin{tabular}{cc}
    \hline
     Model & Avg. Accuracy $\%$  \\ \hline
     NNConv+GCN & $67.2 \pm 0.4$ \\
     GCN+GCN & $68.7 \pm 0.5$ \\
     GCN+GAT & $66.2 \pm 0.5$ \\
     NNConv+NNConv & $70.7 \pm 0.4$ \\
     NNConv+GAT & $64.6 \pm 0.7$ \\
     GEN+GAT & $66.5 \pm 0.8$ \\ \hline
\end{tabular}
    \caption{The accuracy of the trained GNNs depending on the choice of convolutional layers.}
    \label{tbl:accuracy}
\end{table}

\section{Conclusions}

To the already intertwining subjects of string theory and low-dimensional topology, machine learning, particularly Graph Neural Networks, is a powerful tool that can help optimize certain computations.
As shown in this paper, a natural investigation framework is the setting of graph-manifolds. These spaces, introduced in Sections \ref{sec:Intro} and \ref{sec:3-manifold}, can be encoded in a graph-like structure that, when altered by a set of operations, produces a structure associated with a space homeomorphic to the original. In principle, given a pair of graphs, it is possible to establish their equivalence in super-polynomial time by repetitive application of the von Neumann moves. Whereas this feat is easily doable for \textit{small} graphs, escalating to more complex structures quickly becomes computationally unfeasible. A Graph Neural Network, as in Section \ref{sec:GNN}, can be trained to perform the job in polynomial time distinguishing homeomorphic pairs even when the graph becomes too big. The structure of the network comprises two convolutional layers, an aggregation layer, and a classifier layer that converts the graph data into a binary output. To train the GNNs, we created a dataset of pairs of homeomorphic graphs and trained the network on it, achieving for the best model an accuracy of $70.7\%$. We emphasize again that this value has been reached at the end of the training time, and thus a longer training time with a larger dataset is expected to improve the accuracy of the trained network. 

An obvious extension to the current work is to consider a reinforcement learning approach. Contrary to the supervised learning approach taken here, reinforcement learning does not merely ask if there exists a path in configuration space connecting the two inputs, but it attempts to determine the path itself; of the exponentially large collection of possible moves, the network learns heuristics for which sequences of moves to apply to arrive at the desired configuration. The output is a sequence of von Neumann moves, constituting a homeomorphism between the two graph-manifolds, which means that it can be checked ``by-hand'' in polynomial time; thus providing a mathematically rigorous result. For a summary of some of the successes of reinforcement learning in physics and mathematics, see \cite{Gukov:2024buj}. This approach had success with the genus-zero acyclic graph-manifolds that were studied in \cite{Ri:2023xcn}, and thus it is not unexpected that it would be similarly efficacious here.

Continuing along the direction taken in this paper, many more analyses can be pursued. A natural extension is to study Kirby diagrams for generic three-manifolds and ask about their equivalence under the Kirby moves, as described in Section \ref{sec:Intro}. It is important to note that a Kirby diagram is generally \emph{not} a graph, and thus the well-developed GNN techniques that have been developed for the study of graphs cannot be directly applied. While Kirby diagrams can admit a presentation via decorated graphs, it is necessary to develop an encoder from the generic structure to a graph embedding as part of the convolutional layer of the GNN, and to further take account of the possible non-uniqueness of this embedding. 

Furthermore, there exists a generalization of Kirby diagrams to smooth four-manifolds, and a collection of Kirby moves that relate diffeomorphic four-manifolds. GNN approaches can then again be used to search for interesting examples of homeomorphic topological four-manifolds with different choices of smooth structures that are or are not diffeomorphic to each other. In this way, GNN (or more generically Geometric Deep Learning \cite{DBLP:journals/corr/abs-2104-13478}) techniques may highlight interesting corners of the wild world of four-manifolds. See \cite{Gukov:2024buj} for a discussion of some machine learning approaches to the study of smooth structures in four dimensions. 

Finally, one of the principle motivations for this work was the development of $q$-series invariants of three-manifolds which are related to the counting of BPS operators in certain 3d $\mathcal{N}=2$ supersymmetric quantum field theories \cite{Gukov:2017kmk,Gukov:2016gkn}. The quantum field theory of interest is obtained by compactifying one of the 6d $(2,0)$ superconformal field theories on the three-manifold whose invariants we want to determine; only for (weakly-negative-definite) graph-manifolds, such as the ones studied in this paper, are concrete expressions for these $q$-series invariants known. There are a variety of interesting questions connecting the three-manifolds with the spectrum of BPS operators of the quantum field theory for which a machine learning approach may be useful for the generation of conjectures that can later be verified mathematically. A simple example: instead of predicting whether two plumbing graphs correspond to homeomorphic three-manifolds, can we use supervised learning to predict (the strictly weaker condition of) when they have the same first $N$ coefficients in the $q$-series.\footnote{Generalizations and refinements of the $q$-series, such as those in \cite{Ri:2022bxf}, may also be fruitfully explored in this way. Similarly, some relations between the $q$-series invariants are known \cite{Cheng:2023row}; a GNN approach may help to understand this additional structure in the space of graph-manifolds.} Said differently: can we predict the number of protected operators in a quantum field theory by studying the graph of the three-manifold that engineers it? Some steps in the direction of machine learning involving $q$-series invariants have recently been explored in \cite{Gukov:2024opc}.

More generally, transforming super-polynomial classification problems into a neural network formulation can lead to a polynomial time answer at the cost of accuracy. Thus, a well-trained neural network can function as a useful tool to determine interesting correspondences, such as certain pairs of plumbed three-manifolds with particular graph structures that appear to be (not) homeomorphic, and to which analytical toolkits can subsequently be turned.

\section*{Acknowledgements}
The authors thank the Simons Summer Workshop 2024 for hospitality during an important stage of this work. The authors acknowledge support from DESY (Hamburg, Germany), a member of the Helmholtz Association HGF. This work was partially supported by the Deutsche Forschungsgemeinschaft under Germany’s Excellence Strategy -- EXC 2121 “Quantum Universe” -- 390833306 and the Collaborative Research Center -- SFB 1624 “Higher Structures, Moduli Spaces, and Integrability” -- 506632645.

\appendix

\section{Algorithms}\label{app:Algorithms}

For the supervised learning analysis in this paper, it was necessary to have a database of pairs of plumbing graphs where it is known whether or not the associated graph-manifolds are homeomorphic. In this appendix, we briefly summarize the procedure by which these datasets were created. We refer the reader to \href{https://github.com/Lollo0900/Plumbed_3-Manifolds}{\color{blue}$\text{GitHub}^{\includegraphics[scale=0.07]{github-mark.pdf}}$} for the full details.

Three main routines were used in this work: \texttt{EquivPair}, \texttt{InEquivPair}, and \texttt{TweakPair}.\footnote{For consistency, we have used the same routine naming conventions as in \cite{Ri:2023xcn}.} Each routine takes advantage of the generation of a random plumbing graph. For the datasets used in this work, these graphs have a number of vertices between $1$ and $25$, and the triple $(e_i, g_i, r_i)$ associated with each vertex (see equation \eqref{eqn:XXX}) is uniformly distributed in the following intervals
\begin{equation}
    e_i \in [-20,20] \,, \quad g_i \in [-4,4] \,, \quad r_i \in [0,-2] \,.
\end{equation}
The edge labels are distributed uniformly between $\pm 1$. 

Specifically, \texttt{EquivPair} starts from such a random graph and copies it. Separately, on the original and its copy, it then applies on a random vertex, if possible, one of the von Neumann moves of Section \ref{sec:3-manifold}. This procedure is repeated $N_\text{max}$ number of times, after which the function outputs the two resulting graphs, as well as a label $1$ indicating that the pair are homeomorphic.

The $\texttt{InEquivPair}$ routine works similarly to $\texttt{EquivPair}$, but instead of starting with a graph and its duplicate, it starts with two distinct randomly generated graphs.\footnote{The probability of randomly generating two homeomorphic graphs with the chosen set parameters is negligible.} After $N_\text{max}$ randomly applied von Neumann moves, the two graphs are outputted together with label $0$, denoting a non-homeomorphic pair.

Lastly, the $\texttt{TweakPair}$ algorithm starts from a random graph and its copy and then works like $\texttt{EquivPair}$. Differently from the latter, it modifies the values of $e_i$ of a random vertex in the copied graph by an integer in the interval $[-3,3]$, thus generating non-homeomorphic pairs and outputting a $0$ label. 

For this paper, the $N_\text{max}$ parameter is chosen consistently to be $N_\text{max}=60$.

\bibliography{bibliography}{}
\end{document}